\documentclass[8pt,journal,compsoc,onecolumn]{IEEEtran}

\pdfoutput=1

\usepackage{fancyhdr,amsmath,amsthm,amssymb,url,array,textcomp,listings,hyperref,xcolor,colortbl,float,gensymb,longtable,supertabular,multicol,placeins,booktabs,lineno,graphicx,caption,subfig}

\graphicspath{{./Figures/}}
\DeclareGraphicsExtensions{.pdf,.png,.jpg,.eps}

\usepackage[printwatermark]{xwatermark}
\newwatermark*[allpages,color=red,angle=0,scale=1,xpos=0,ypos=380pt,fontsize=6pt]{
The content of this paper was published in IET Biometrics, 2018. This ArXiv version is from before the peer review. Please, cite the following paper:
\\Ziga Emersic, Luka Lan Gabriel, Vitomir Struc, Peter Peer: "Convolutional Encoder-Decoder Networks for Pixel-Wise Ear Detection and Segmentation", IET Biometrics, 2018
}

\begin{document}
\title{Pixel-wise Ear Detection with Convolutional Encoder-Decoder Networks}

\author{\v{Z}iga Emer\v{s}i\v{c} $^{1}$, Luka Lan Gabriel $^{2}$, Vitomir \v{S}truc $^{3}$ and Peter Peer $^{1}$%
\thanks{ $^{1}$ \quad Faculty of Computer and Information Science, University of Ljubljana, Ve\v{c}na pot 113, SI-1000 Ljubljana; \{ziga.emersic, peter.peer\}@fri.uni-lj.si}%
\thanks{$^{2}$ \quad KTH Royal Institute of Technology, Kungl Tekniska Högskolan, SE-100 44 Stockholm; llga@kth.se}%
\thanks{$^{3}$ \quad Faculty of Electrical Engineering, University of Ljubljana, Tr\v{z}a\v{s}ka 25, SI-1000 Ljubljana; vitomir.struc@fe.uni-lj.si}%
\thanks{} 
\thanks{Submitted for publication.}%
}

\IEEEtitleabstractindextext{%
\begin{abstract}
Object detection and segmentation represents the basis for many tasks in computer and machine vision. In biometric recognition systems the detection of the region-of-interest (ROI) is one of the most crucial steps in the overall processing pipeline, significantly impacting the performance of the entire recognition system. Existing approaches to ear detection, for example, are commonly susceptible to the presence of severe occlusions, ear accessories or variable illumination conditions and often deteriorate in their performance if applied on ear images captured in unconstrained settings. To address these shortcomings, we present in this paper a novel ear detection technique based on convolutional encoder-decoder networks (CEDs). For our technique, we formulate the problem of ear detection as a two-class segmentation problem and train a convolutional encoder-decoder network based on the SegNet architecture to distinguish between image-pixels belonging to either the ear or the non-ear class. The output of the network is then post-processed to further refine the segmentation result and return the final locations of the ears in the input image. Different from competing techniques from the literature, our approach does not simply return a bounding box around the detected ear, but provides detailed, pixel-wise information about the location of the ears in the image. Our experiments on a dataset gathered from the web (a.k.a. in the wild) show that the proposed technique ensures good detection results in the presence of various covariate factors and significantly outperforms the existing state-of-the-art.
\end{abstract}

}

\maketitle

\section{Introduction}\label{sec:introduction}

Ear recognition has gained on popularity in the recent years due to the numerous application possibilities in forensics, security and surveillance. 
However, despite its popularity, only a limited number of fully automatic techniques has been proposed and presented in the literature. Many recent surveys on ear recognition ascribe this fact to the lack of efficient ear detection techniques capable of determining the location of the ear(s) in the input images, which are a key component of automatic ear recognition systems~\cite{abaza2013survey,anika_survey,earrecog}. In fact, Emer\v{s}i\v{c} et al.~\cite{earrecog} argue that the absence of automatic ear detection approaches is one the most important factors hindering a wider deployment of ear recognition technology.

While there has been progress in the area of ear detection in recent years (see, \cite{anika_survey} for a comprehensive survey on existing ear detection approaches), most of the existing work is limited to laboratory-like settings and controlled image acquisition conditions, where the appearance variability of ear images is limited and not representative of real-world imaging conditions. In unconstrained setting, the problem of ear detection is extremely challenging, as ears can appear in various shapes, sizes, and colors, they can be occluded by hair strains or accessories and the imaging conditions can vary in terms of illumination and viewing angles. In these settings, ear detection is still an unsolved problem and no widely
adopted solution has been (to the best of our knowledge) proposed yet in the literature.  

In this paper, we try to address this gap and introduce a novel ear detection approach based on convolutional neural networks (CNNs). We pose the ear detection problem as a two-class image segmentation task, where the goal is to assign each image pixel to either the ear or to the non-ear class. We train a convolutional encoder-decoder network (CED) based on the SegNet~\cite{segnet, segnet2} architecture to classify the image pixels in one of the two classes and use the trained network to generate an initial segmentation result from the input image. We then further refine the result through a postprocessing procedure that takes anthropometric assumptions into account and removes spurious image region from the final output. Different from the existing solutions to ear detection in the literature, which typically return only a bounding rectangle or ellipse for each detected ear in the image, our approach provides information about the locations of the ears at the pixel level. Such information is useful for ear recognition systems as it allows to exclude background pixels from the feature extraction and classification stages, which is not possible with standard detection techniques.

We evaluate our pixel-wise ear detection approach based on convolution encoder-decoder networks (PED-CED) in experiments on the AWE dataset~\cite{earrecog}. The AWE dataset is a recent dataset of ear images gathered from the web with the goal of studying unconstrained ear recognition technology and is in our opinion the most challenging datasets available in this area. In contrast to other datasets typically used to assess ear detection performance, images from the AWE dataset are not limited to perfect profile face images (i.e., faces rotated close to $90^\circ$ yaw angles), but feature significant variations in yaw angles as well. We present a detailed experimental analysis of our proposed PED-CED approach and study the influence of various covariates on the detection performance. We also report comparative experiments with state-of-the-art recognition techniques from the literature. Our results indicate that the PED-CNN technique is a viable option for ear detection in unconstrained settings and provides competitive detection performance even in the presence of different covariate factors despite the fact that only a limited number of images was available for training.

To summarize, we make the following contributions in this paper:
\begin{itemize}
\item we present a novel ear detection technique based on convolutional neural networks (CNNs) that works well on image data captured in completely unconstrained settings and returns a pixel-wise segmentation of the ear area from the given input image,
\item we provide a detailed analysis of the proposed techniques with respect to various covariates and identify open problems that need to be addressed to further improve its performance, and
\item we report comparative results with state-of-the-art ear detection techniques from the literature.
\end{itemize}

\section{Related Work}\label{Sec: Related work}
In this section we survey the most important techniques for ear detection with the goal of positioning our work within the area. For a more comprehensive review on existing ear detection approaches (from 2D as well as 3D imagery), the reader is referred to  recent surveys on this topic~\cite{anika_survey, prakash_gupta_book}.

It needs to be noted that no standard benchmarks and evaluation methodology exist for ear detection technology, which makes it difficult to compare existing approaches among each other. Authors typically report different performance metrics and rely on self compiled evaluation protocols in their experiments. The reader should therefore avoid making direct comparisons between the  detection accuracies presented in the remainder of this section. Furthermore, since face detection is commonly assumed to have been run on the images before ear detection is performed, the term ear detection is  typically used  interchangeably with ear localization or even ear enrollment, though these two terms in general describe a somewhat narrower problem that already assumes that one (or two) ear(s) are present in the input image.

In~\cite{a38} the authors propose an ear enrollment algorithm that fits an ellipse to the ear using the Hough
Transform. The approach is sufficiently tolerant to noise and occlusions and achieves a 91\% enrollment success rate on the UND dataset~\cite{und} and 100\% on XM2VTS when no occlusions are present. The authors do not explicitly state what constitutes a successfull enrollment attempt. 

In~\cite{a30} the Canny edge detector is used to extract edges from ear images and the ears outer helix curves are used as features for the localization process. On the IITK dataset~\cite{prakash_gupta_book} the authors report the localization accuracy of 93.34\%, where the accuracy is defined as:
\begin{equation}
\text{accuracy} = \frac{\text{\# of correct detections/localizations}}{\text{\# of all annotated ears}},
\label{eq_1}
\end{equation}
In another work using the Canny edge detector~\cite{a29}, the authors report the localization accuracy of 98.05\% on the USTB~\cite{ustb} and 97.05\% on the Carreira-Perpinan dataset~\cite{carreira}, but similar to~\cite{a30} do not provide information on how a correct ear detection/localization is defined (i.e., the nominator of Eq.~\ref{eq_1}). 

In the work of~\cite{islam}, a cascaded-AdaBoost-based ear detection approach is proposed. The authors report the detection rate of 100\% with the false positive rate of $5\times10^{-6}$ on 203 profile images from the UND dataset~\cite{und}. Again no strict criterion is given by the authors about the process of establishing the detection and false positive rates, though it is suggested that the results were checked manually.

Another approach to ear detection based on the distance transform and template matching is proposed in~\cite{ear_loc}. The authors report the detection accuracy (using Eq.~(\ref{eq_1})) of 95.2\% on the IIT Kanpur ear database. The authors define a correct detection as one that exhibits a sufficient level (i.e., above some predefined threshold) of similarity with generic ear template. 

In~\cite{prakash2009} the connected component analysis of a graph constructed using the edge map of the image and is evaluated on a data set consisting of 2361 side face images from IITK dataset~\cite{prakash_gupta_book}. The authors fit rectangular regions to the ear images and achieve the detection accuracy of 95.88\% on 490 test images and the detection accuracy of 94.73\% on another test set of 801 images, when at most 15\% more pixels are detected around the annotated ground truth.

In another paper~\cite{prakash2009skin} the same authors approach the ear detection problem by segmenting skin-colored regions. Using 150 side face images of the IITK dataset~\cite{prakash_gupta_book} the approach achieves 94\% detection accuracy. The accuracy is again measured through Eq.~(\ref{eq_1}) and similarly to~\cite{ear_loc} the correctness of the detection is established based on the level of similarity of the detected region and a generic ear template.

Haar features arranged in a cascaded Adaboost classifier, better known as Viola-Jones~\cite{viola-jones}, are used in~\cite{a33} for ear detection. The authors manually annotate the UND-F~\cite{und}, UMIST~\cite{umist}, WV HTF~\cite{a33} and USTB~\cite{ustb} datasets with rectangles around ears and use the annotated data for training and testing. The authors achieve ~95\% detection accuracy on the combined images and 88.72\% on the UND-F dataset. Similar to our technique, this approach is also capable of handling a wide variety of image variability and operating in real-time.

The authors of~\cite{a39} propose an ear enrollment technique using the image ray transform, which highlights the tubular structures of ear. Using 252 images from the XM2VTS~\cite{xm2vts} dataset the authors achieve a 99.6\% enrollment rate and consider an image as successfully enrolled if after the enrollment/localization process, the entire ear is contained in the localized image area.

The approach presented in~\cite{prakash2012} makes use of the edge map of the side face images. An edge connectivity graph build on top of the edge map serves as the basis for ear candidate calculation. The detection performance is evaluated on the IITK, UND-E and UND-J2 datasets, achieving 99.25\% accuracy on IITK. As suggested by the authors, the detection accuracy is defined by Eq.~(\ref{eq_1}), but no criterion defining a correct detection is given by the authors.

The HEARD~\cite{heard} ear detection method  is based on three main features of a human ear: the height-to-width ratio of the ear, the area-to-perimeter ratio of the ear, and the fact that the ear's outline is the most rounded outline on the side of a human face. To avoid occlusions caused by hair and earrings the method looks for the inner part of the ear instead of the outer part. The authors use the UND~\cite{und}, CLV~\cite{cvl} and IITK~\cite{prakash_gupta_book} datasets. The method is able to detect 98\% of ears in the UND-E dataset~\cite{und}. However, no information is given by the authors on how the reported detection accuracy is calculated. The detection time for a single image is 2.48 seconds, making the method significantly slower than our PED-CED approach.

The ear detection algorithm proposed in~\cite{robust_localization} uses texture and depth images for localizing ears in both images of profile faces and images taken with at different camera angles. Details on the ear surface and edge information are used for determining the ear outline in an image. The algorithm utilizes the fact that the surface of the outer ear has a delicate structure with high local curvature. The ear detection procedure returns an enclosing rectangle of the best ear candidate. The detection rate of this algorithm is 99\%. A detection was considered successful when the overlap $O$ between the ground truth pixels $G$ (i.e., the annotated area) and the pixels in the detected region $R$ is at least 50\%. The overlap $O$ is calculated by the following equation:
\begin{equation}
O=\frac{2\left | G\bigcap R \right |}{\left | G \right | + \left | R \right |},
\label{eq_robust}
\end{equation}
where $\bigcap$ stands for the intersection operator and $+$ for the union.

In~\cite{ganesh} the authors present a method called Entropic Binary Particle Swarm Optimization (EBPSO) which generates an entropy map, the highest value of which is used to detect the ear in the given face image. Additionally background subtraction is performed. The authors evaluate the detection accuracy using the CMU PIE~\cite{cmu-pie}, Pointing Head Pose~\cite{php}, FERET~\cite{feret} and UMIST~\cite{umist} datasets. On FERET the authors report the detection rate of 94.96\%, where the detection rate is defined by Eq.~(\ref{eq_1}) and a detection attempt is considered successful if at least part of the ear is contained in the detected area.

In~\cite{prajwal} the authors report results on the same datasets as~\cite{ganesh}, with the addition of FEI~\cite{fei}. The proposed approach uses the entropy-Hough transform for enhancing the performance of the ear detection system. A combination of a hybrid ear localizer and an ellipsoid ear classifier is used to predict locations of ears. The authors achieve 100\% detection rate on the UMIST~\cite{umist} and FEI~\cite{fei} datasets and 73.95\% on FERET. 
The detection rate is computed with Eq.~(\ref{eq_1}) and a detection attempt is considered successful if the center of the detected region is close enough to the center of the annotated ground truth (i.e., the distance is below some threshold) and the detected area contains the entire ear.

The authors of~\cite{auto_ear} present a new scheme for automatic ear localization: template matching based on the modified Hausdorff distance. The benefit of this technique is that it does not depend on pixel intensities and that the template incorporates various ear shapes. Thus, this approach is invariant to illumination, pose, shape and occlusion in profile face images. The detection accuracy of the technique was tested on two datasets, i.e., the CVL face database~\cite{cvl} and the UND-E database~\cite{und}, on which accuracies of 91\% and 94.54\% were obtained, respectively. The accuracy is calculated by Eq.~(\ref{eq_1}), but no criteria for a correct detection are reported.

A summary of the surveyed work is presented in Table~\ref{table_comparison}. Note again that reported accuracies are not directly comparable, as different datasets, performance metrics and evaluation protocols were used by the authors.

\begin{table*}[!htb]
\centering
\begin{tabular}{lllll}
\hline
Year & Detection Method & Dataset & \begin{tabular}[c]{@{}l@{}}\# test\\ images\end{tabular} & Accuracy* [\%] \\ \hline
2007 & Hough Transform~\cite{a38} &  UND~\cite{und} & 942 & 91 \\
	& Canny Edge Detector~\cite{a30} & IITK~\cite{prakash_gupta_book} & 700 & 93.34 \\
2008 
 	& Canny Edge Detector and Line Tracing~\cite{a29} & USTB~\cite{ustb} & 308 & 98.05 \\
 	& Adaboost~\cite{islam} & UND~\cite{und} & 203 & 100 \\
 	& Distance Transform and Template Matching~\cite{ear_loc}& IITK~\cite{prakash_gupta_book} & 150 & 95.2 \\
2009 & Connected component~\cite{prakash2009} & IITK~\cite{prakash_gupta_book} & 490 & 95.88 \\
	& Skin-Color~\cite{prakash2009skin} & IITK~\cite{prakash_gupta_book} & 150 & 94 \\
2010 & Viola-Jones~\cite{a33} & UND-F~\cite{und} & 940 & 88.72 \\
	& Ray Transform~\cite{a39} & XM2VTS~\cite{xm2vts} & 252 & 98.4 \\
2012 & Skin Color and Graph Matching~\cite{prakash2012} & IITK~\cite{prakash_gupta_book} & 1780 & 99.25 \\ 
	& HEARD~\cite{heard} & UND-E~\cite{und} & 200 & 98 \\
2013 & Feature Level Fusion and Context Information~\cite{robust_localization} & UND-J2~\cite{und} & 1776 & 99 \\
2014 & EBPSO~\cite{ganesh} & FERET~\cite{feret} & 258  & 94.96 \\
2015 & Entropy Hough transform~\cite{prajwal} & UMIST~\cite{umist} & 564 & 100 \\
2016 & Modified Hausdorff Distance~\cite{auto_ear} & UND-E~\cite{und} & 464 & 94.54 \\
\hline
\end{tabular}
\caption{Summary of the surveyed work on ear detection using 2D images. While some works evaluate their methods on multiple datasets, we report only results for the dataset with the highest detection accuracy. * Not all accuracies were calculated using the same equation. Furthermore, different datasets and protocols were used, so conclusions about the relative performance of the listed techniques should be avoided.}
\label{table_comparison}
\end{table*}

\noindent

\section{Pixel-wise Ear Detection with CEDs}
In this section we present our Pixel-wise Ear Detection technique based on Convolutional Encoder-Decoder networks (PED-CED). We
start the section with a high-level overview of the proposed technique and then proceed with a description of the
individual building blocks.

\subsection{Overview}

Our PED-CED ear detection technique is based on the assumption that a single face is present in the input image and that our goal is to detect all visible
ears present. This is a reasonable assumption, as the input image is typically subjected to a face detection procedure prior
to ear detection. No other assumptions regarding the input images are made, they are free to vary in terms of appearance, imaging conditions and alike.

\begin{figure*}[b]
\centering
\includegraphics[width=1.0\textwidth]{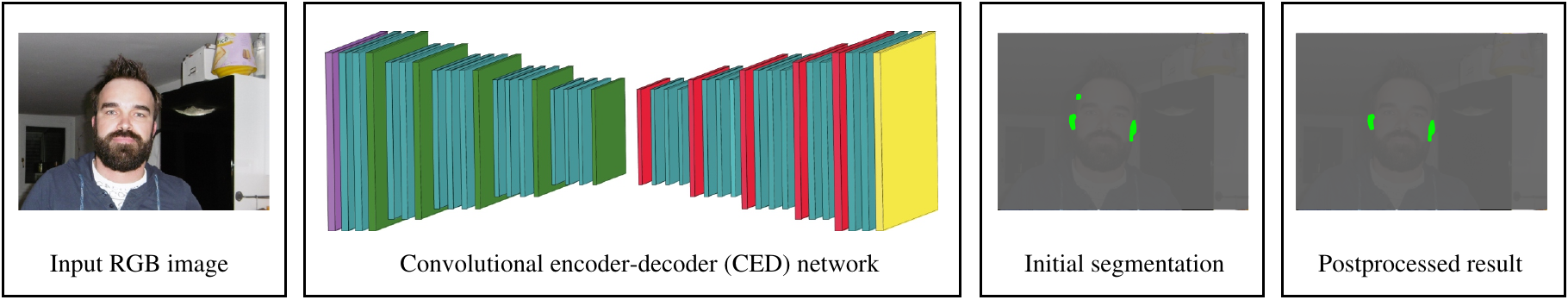}
\caption{Overview of the PED-CED ear detection approach. Ear detection is posed as a segmentation problem and solved using a convolutional encoder-decoder network (i.e., the segmentation network). The segmentation network takes an RGB-image (containing a face) as input and returns ear candidate regions as a result. The segmentation output is then postprocessed and (at most) the two largest areas are retained (Figure is best viewed in color).}
\label{figure_segnet_overview}
\end{figure*}

To locate the ears in the  image, we rely on the recently proposed convolutional encoder-decoder (CED) network, termed SegNet~\cite{segnet, segnet2}. The network
allows us to label each pixel in the input image with an ear or non-ear label and identify ear candidate locations in the image. Because the segmentation procedure sometimes
returns spurious labels, we postprocess the segmentation results and retain (at most) the two largest ear regions. This step is in accordance
with our assumption that a single face is present in the input image and hence at most two image regions can be found by the detection procedure. The entire pipeline is illustrated in Figure~\ref{figure_segnet_overview}.

A detailed description of the individual components of our PED-CED approach is given in the following sections.

\subsection{The segmentation network}

As emphasized in the previous section, our segmentation network is based on the SegNet model architecture, which was recently proposed for robust semantic pixel-wise labeling~\cite{segnet,segnet2}. The architecture was modified for this work, enabling robust ear segmentation from unconstrained images. 

SegNet~\cite{segnet} consists of a sequence of nonlinear processing layers (encoders) and a corresponding set of decoders with a pixel-wise classifier. As shown in the second block of Figure~\ref{figure_segnet_overview}, a single encoder consists of several convolutional layers, a batch normalization layer and a rectified-linear-unit ReLU layer (shown in blue), followed by a  max-pooling and sub-sampling layer (shown in green). The goal of the encoders is to produce low-resolution feature maps that compress the semantic information in the input image and can be fed to the sequence of decoders for upsampling. A key feature of the SegNet architecture is the use of max-pooling indices from the encoders in the corresponding decoders, which enables non-linear up-sampling of the low-resolution feature maps. With this procedure high frequency details in the segmented images are retained and the total number of parameters in the decoders is reduced. Similar to the encoders, each decoder is also composed of several convolutional layers, a batch normalization layer and a ReLU layer (shown in blue in Figure~\ref{figure_segnet_overview}) followed by an upsampling layer (shown in red). The final layer of the segmentation network is a pixel-wise softmax layer, which in our case, assigns each pixel a label corresponding to one of two classes (i.e., ear or non-ear). 

The CED network encoder used in this work represent the popular VGG-16 architecturee~\cite{simonyan2014very} composed of  13 convolutional layers interspersed with max pooling layers. The decoder has a similar architecture but instead of max pooling layers contains upsampling layers that upsample the feature maps generated by the encoder to a larger size. The whole architecture is summarized in Table~\ref{tab_architecture}. Here, information on the type of each layer and size of outputs is provided.
\begin{table}[htb]
\centering
\begin{tabular}{ccc}
Layer Number & Type of Layer & Number of Filters\\ \hline
- & data/input & \\ \hline
1, 2 & convolutional & 64 \\ \hline
- & max pooling &  \\ \hline
3, 4 & convolutional & 128 \\ \hline
- & max pooling &   \\ \hline
5, 6, 7 & convolutional & 256 \\ \hline
- & max pooling &   \\ \hline
8, 9, 10 & convolutional & 512 \\ \hline
- & max pooling &   \\ \hline
11, 12, 13 & convolutional & 512 \\ \hline
- & max pooling &   \\ \hline
- & upsample &   \\ \hline
14, 15, 16 & convolutional & 512 \\ \hline
- & upsample &   \\ \hline
17, 18 & convolutional & 512 \\ \hline
19 & convolutional & 256 \\ \hline
- & upsample &   \\ \hline
20, 21 & convolutional & 256 \\ \hline
22 & convolutional & 128 \\ \hline
- & upsample &   \\ \hline
23 & convolutional & 128 \\ \hline
24 & convolutional & 64 \\ \hline
- & upsample &   \\ \hline
25 & convolutional & 64 \\ \hline
26 & convolutional & 2 \\ \hline
- & softmax & \\
\end{tabular}
\caption{High-level summary of the layers used in our CED architecture. A convolutional layer is always followed by a BN and a ReLU layer.}
\label{tab_architecture}
\end{table}

To train the segmentation network we use 750 uncropped images from the AWE dataset~\cite{earrecog}, which were gathered from the web for the goal of studying ear recognition technology in unconstrained settings and therefore exhibit a high-degree of variability. We use stochastic gradient descent for the training procedure and set the learning rate to the value of 0.0001, the momentum to 0.9 and the weight decay to 0.005~\cite{learning_rate},~\cite{momentum},~\cite{weight_decay}. We use the publicly available Caffe implementation of the SegNet model~\footnote{https://github.com/alexgkendall/caffe-segnet} to initialize our segmentation network, but modify the last convolutional and softmax layer. Here, we set the number of outputs of the
last convolutional layer to 2 (ear and non-ear) and calculate new class weights that we apply to the softmax layer to ensure stable network training. As advocated in~\cite{segnet, segnet2}, we also use median frequency balancing~\cite{freq_balanc} to compute the class weights in our cross-entropy loss, which compensates for the fact that pixels from the ear class cover only a small portion of the input image, while the the remaining pixels belong to the non-ear class. Without frequency balancing, the network would likely converge to a trivial solution, where all pixels would be assigned to the dominant (over-represented) non-ear class.

\begin{figure}[htb]
\centering
\includegraphics[width=0.78\columnwidth]{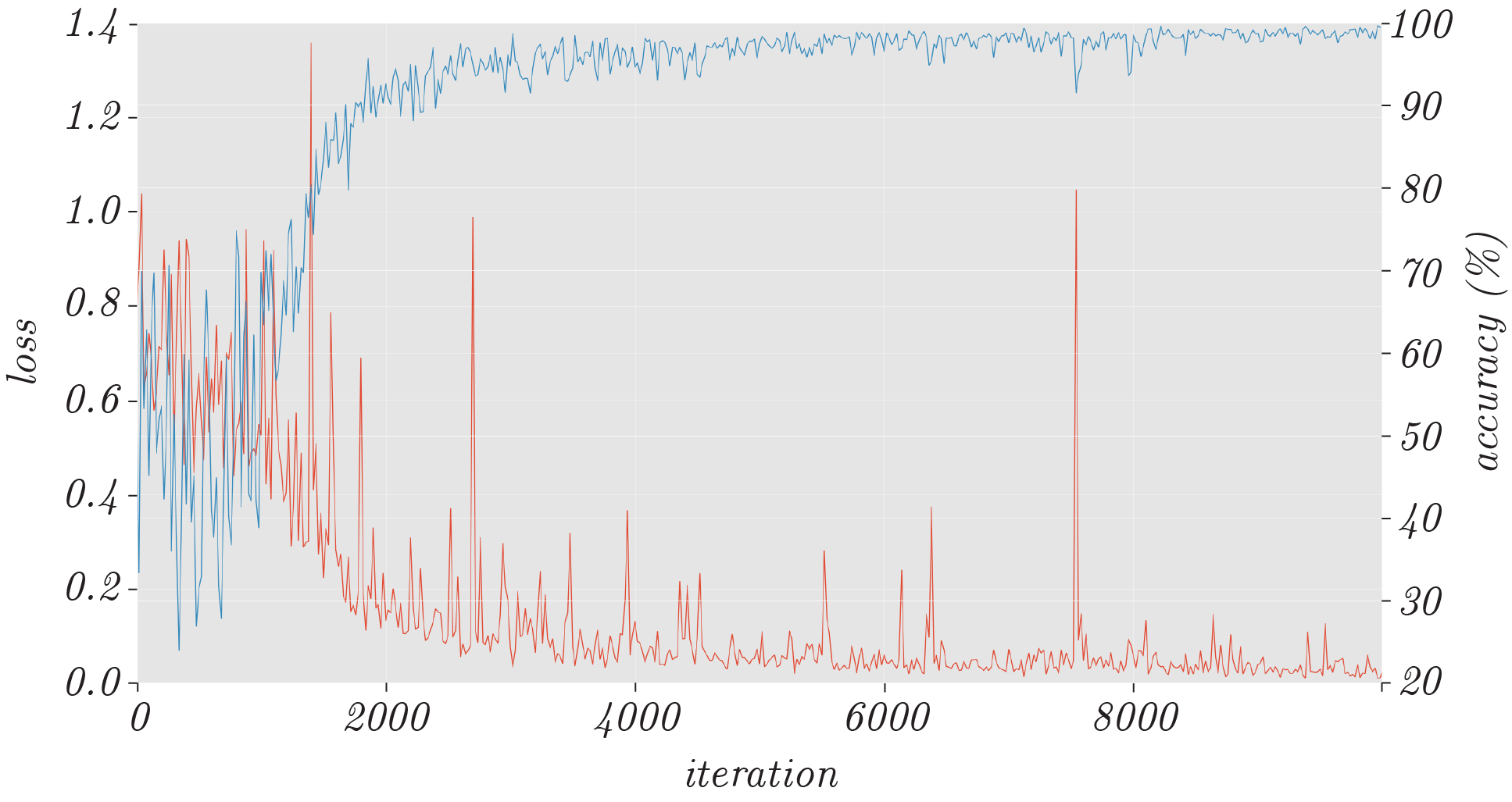}
\caption{Graphical representation of the convergence of the loss (red) and accuracy (blue) during network training. The values were sampled every 20 training iterations.}
\label{figure_training}
\end{figure}

The training is conducted on a graphical processing unit (GPU). All training images are therefore resized to the resolution of 480$\times$360 pixels to reduce the needed graphical memory prior to training. Figure~\ref{figure_training} shows the loss and accuracy values, collected in steps of 20 iterations throughout the course of the training process. We can see that the training converges after approximately 10000 iterations, when  stable loss and accuracy values are reached.

\subsection{Postprocessing}

\begin{figure}[htb]
\centering
\includegraphics[width=0.32\columnwidth]{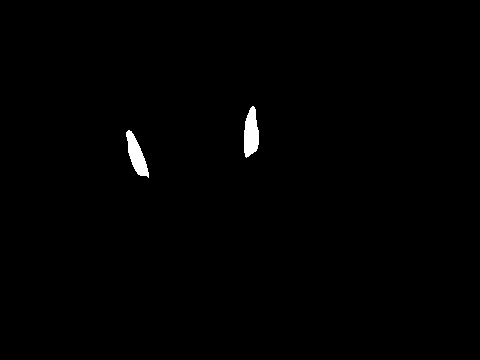}
\includegraphics[width=0.32\columnwidth]{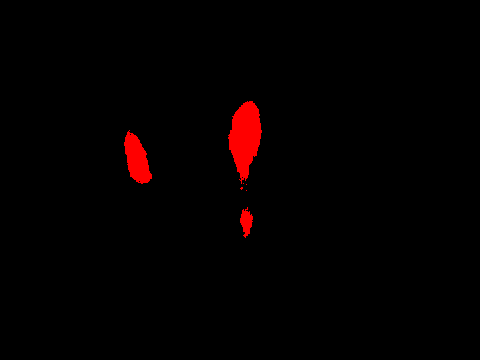}
\includegraphics[width=0.32\columnwidth]{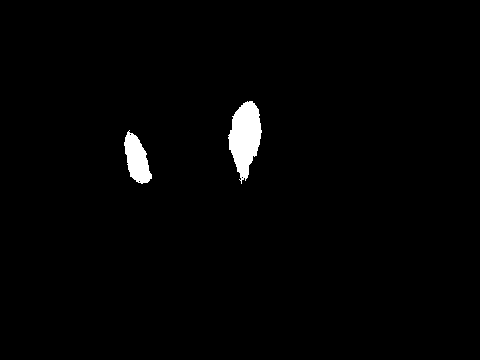}
\caption{An example of the post processing procedure: The first image shows the annotated ground-truth, the second image shows the initial segmentation result, and the third the final post-processed detection areas. The post-processing step retains only the two largest ear regions and discards all other potentially spurious regions.}
\label{figure_post_process}
\end{figure}

Once the segmentation network is trained it can be used to generate initial segmentation results from the input images. However, these results are not always perfect and despite the fact that only images with a single face are expected as input to our PED-CED detection procedure, several ear candidate regions may be present in the initial segmentation results. Since the
only possible correct output of the segmentation network is the detection of one or two regions (corresponding to either one or two visible ears in the image), we apply an additional post-processing step and clean the initial segmentation output. Thus we retain only the two largest regions (or one, if only one was detected) and discard the rest. An example of the procedure is illustrated in Figure~\ref{figure_post_process}. Here, the left image shows the annotated ground truth, the middle images depicts the initial segmentation result and right images shows the final output of our PED-CED detection approach.

This post-processing step removed small, erroneous regions that CNN wrongly classified as ears. However, too-large regions or incorrect regions connected to correct detection were not removed with this procedure. In order to do that additional analysis of shape properties would be needed.

\section{Experimental Setup}
In this section we describe the data, experimental protocols and performance metrics used to assess the efficacy of our PED-CED ear detection approach.


\subsection{Data and experimental protocol}
\begin{figure}[htb]

\centering

\includegraphics[width=0.32\columnwidth]{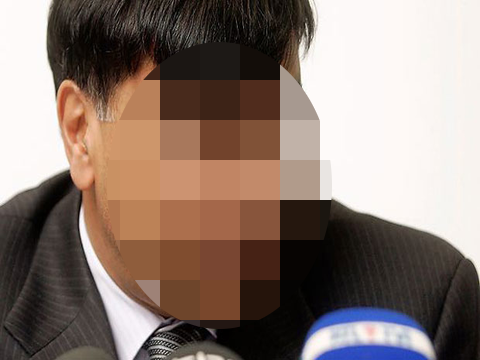}
\includegraphics[width=0.32\columnwidth]{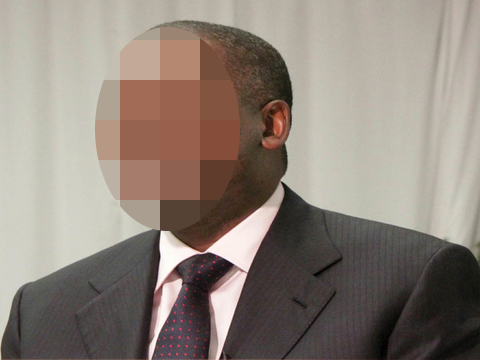}
\includegraphics[width=0.32\columnwidth]{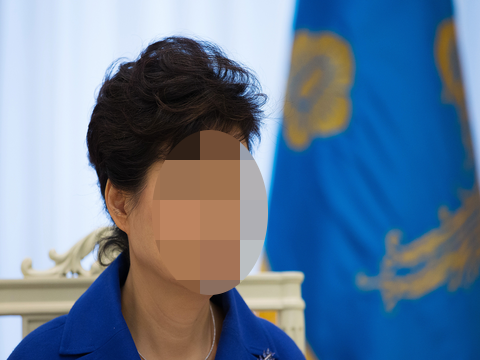}

\vspace{1mm}

\includegraphics[width=0.32\columnwidth]{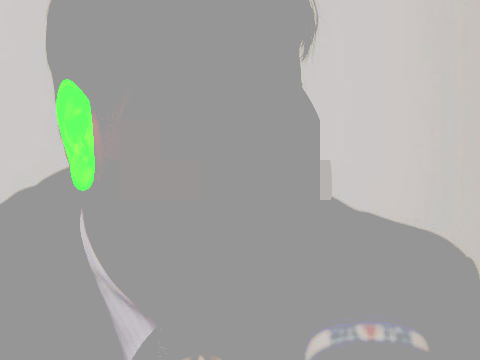}
\includegraphics[width=0.32\columnwidth]{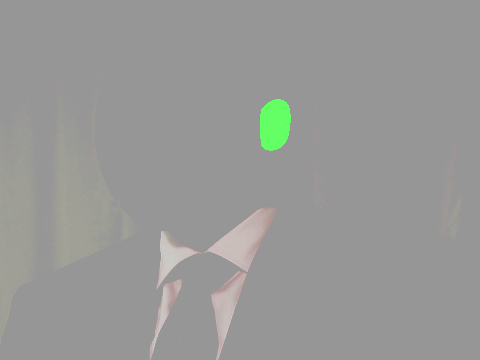}
    \includegraphics[width=0.32\columnwidth]{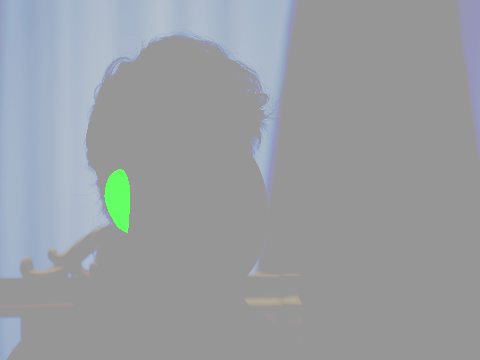}
		\caption{Sample uncropped images from the AWE dataset. The top row shows the input images and the bottom row shows the annotated ear locations. The ears are annotated at the pixel level for all 1000 images of the AWE dataset.  (In the images given here, faces were pixelated in order to guarantee anonymity.)}
		\label{figure:aweOriginalComparison}
\end{figure}

The dataset used in our experiments comprised the original (uncropped) images of the Annotated Web Ears (AWE) dataset~\cite{earrecog}. The cropped version of the dataset together with the evaluation toolbox is freely available~\footnote{http://awe.fri.uni-lj.si}. The dataset contains a total of 1,000 annotated images from 100 distinct subject, with 10 images per subject. All images from the dataset were gathered from the web using a semi-automatic procedure and were labeled according to yaw, roll and pitch angles, ear occlusion, presence of accessories, ethnicity, gender and identity. The dataset has already been used in~\cite{erk1, erk2}. However, new pixel-wise annotations of ear locations had to be created in order to evaluate the segmentation. To this end, a trained annotator manually marked the ear locations in each image at the pixel level and stored the resulting annotations in the form of binary masks for later processing. Figure~\ref{figure:aweOriginalComparison} shows the a few of the original images that we used for our experiments together with the newly created ground truth. Note that the annotations provide more detailed information about the locations of the ears than simple rectangular bounding boxes.

From the available 1000 images in the AWE dataset, we used 750 images for training and 250 images for testing purposes. The training images were used to learn the parameters of the SegNet segmentation network, while the testing images were reserved solely for the final performance evaluation.

\subsection{Performance metrics}\label{Sec: perf metrics}

A typical way of measuring the performance of ear detection technology is to use the detection accuracy, which is typically defined as the ratio between the number of correct detections and the overall number of annotated ear areas. However, as already pointed out in Section~\ref{Sec: Related work}, what is considered a correct detection is usually defined by the authors is not used consistently from paper to paper. Since a single profile face is typically presumed in the input image, the general assumption is that only a single ear needs to detected (localized or enrolled), so false positives are not considered in the reported results, and the decisive criterion is whether the correct ear was found or not.

In this work, we measure the performance of our detection approach by comparing the manually annotated ground-truth locations and the output of our PED-CED approach during testing. We report accuracy values for our approach, which are computed as follows:
\begin{equation}
    Accuracy = \frac{TP + TN}{All},
\label{eq_accuracy}
\end{equation}
where $TP$ stands for the number of true positives, i.e., the number of pixels that are correctly classified as part of an ear, $TN$ stands for the number of true negatives, i.e., the number of pixels that are correctly classified as non-ear pixels, and $All$ denotes the overall number of pixels in the given test image. This accuracy value measures the quality of the segmentation, but is dominated by the non-ear pixels (i.e., the majority class), which commonly cover most of the test image. Thus, our accuracy measure is expected to have large values (close to 1) even if most pixels are classified as belonging to the non-ear class.

The second performance metric used for our experiments is the the Intersection over Union (IoU), which is calculated as follows:
\begin{equation}
	IoU = \frac{TP}{TP + FP + FN},
\end{equation}
where $FP$ and $FN$ denote the number of false positives (i.e., ear pixels classified as non-ear pixels) and number of false negatives (i.e., non-ear pixels classified as ear pixels), respectively. IoU represents the ratio between the number of pixels that are present in both the ground-truth and detected ear areas and the number of pixels in the union of the annotated and detected ear areas. As such it measures it measures the quality (or tightness) of the detection. A value of 1 means that the detected and annotated ear areas overlap perfectly, while a value of less than 1 indicates a worse detection result.

The last two performance metrics used for our experiments are recall and precision, defined as:
\begin{equation}
	Precision = \frac{TP}{TP + FP},
\end{equation}
\begin{equation}
	Recall = \frac{TP}{TP + FN}.
\end{equation}
Precision measures the proportion of correctly detected ear-pixels with respect to the overall number of true ear pixels (i.e., how many detected pixels are relevant), while recall measures the proportion of  correctly detected ear-pixels with respect to the overall number of detected ear pixels (i.e., how many relevant pixels are detected).

\section{Results and Discussion}

We now present the experimental results aimed at demonstrating the merits of our PED-CED ear detection approach. We show comparative results with a state-of-the-art technique from the literature, analyze the impact of various covariates on the performance of our technique and show qualitative examples of our detections.

\subsection{Assessment and comparison to the state-of-the-art}

\begin{figure}[htb]
\centering

\includegraphics[width=0.32\columnwidth]{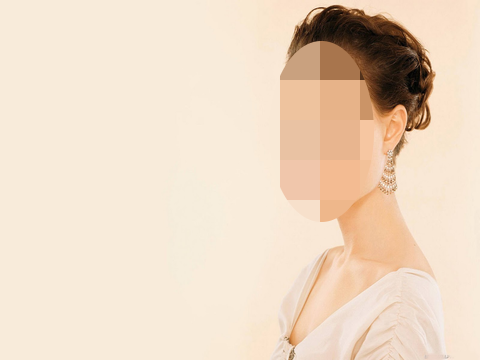}
\includegraphics[width=0.32\columnwidth]{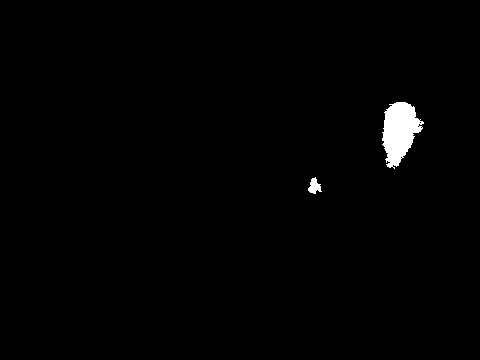}
\includegraphics[width=0.32\columnwidth]{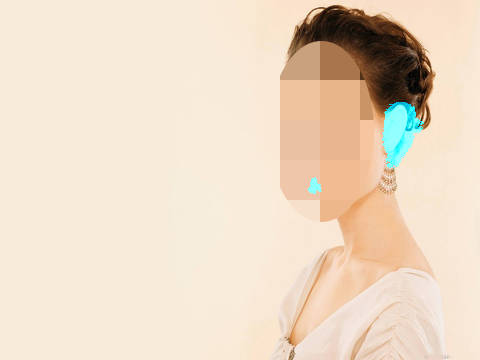}

\vspace{1mm}

\includegraphics[width=0.32\columnwidth]{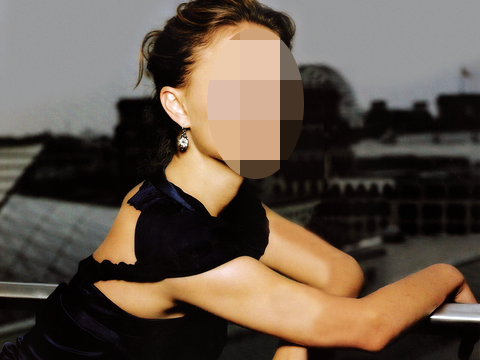}
\includegraphics[width=0.32\columnwidth]{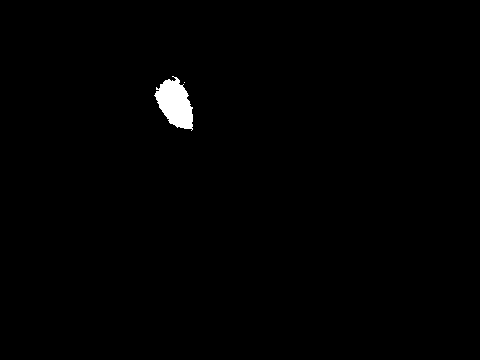}
\includegraphics[width=0.32\columnwidth]{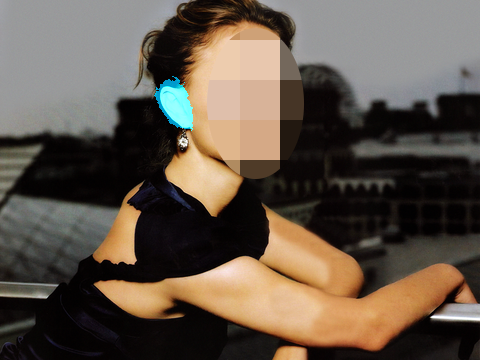}

\caption{Sample detection output of our PED-CED approach. The images in the left column are original test images, the images in the middle represent the detection results, and in the images in the right column show superimposed images composed of the original test image and the detection result. An example of the detection (or better said segmentation) result with some false positives can be seen in the first row: some pixels were incorrectly classified as belonging to the ear class. (In the images given here, faces were pixelated in order to guarantee anonymity.)}
\label{figure_sup}
\end{figure}

The hardware used for experimentation was a desktop PC with Intel(R) Core(TM) i7-6700K CPU with 32GB system memory and an Nvidia GeForce GTX 980 Ti GPU with 6GB of video memory running Ubuntu 16.04 LTS.  On this hardware the training of our segmentation network took 64 minutes and was completed with the network parameters converging after approximately 10000 iterations. The average time for the PED-CED detection procedure on images rescaled to a resolution of $480\times360$ was 87.5 milliseconds, which is orders of magnitude faster from what was reported in~\cite{heard} (2.48 seconds) and also faster than the Haar-based approach from~\cite{a33} - 178.0 milliseconds with our configuration that uses left and right ear cascades. A couple of qualitative results of our detection procedure are shown in Figure~\ref{figure_sup}.

Our PED-CED procedure was run on the entire test set of 250  AWE images and all performance metrics (presented in Section~\ref{Sec: perf metrics}) were computed for each test. The average values of each performance metric together with the corresponding standard deviations are shown in Table~\ref{table_results}. Here, we also include results for the Haar-based ear detector~\cite{a33} based on the object detection procedure proposed by Viola and Jones~\cite{viola-jones} to put our technique into perspective and show comparative results with a state-of-the-art technique from the literature. For the Haar-based ear detector we used the detection cascades for the left and right ear that ship with OpenCV~\cite{lienhart2002extended}. The \texttt{scaleFactor} parameter that specifies how much the image size is reduced at each image scale was set to $1.05$. The \texttt{minNeighbors} parameter that specifies how many neighbors each candidate rectangle should have was set to $5$. Other parameters were left at default values. To be able to compare both techniques on equal footing, we calculate bounding rectangles for our ground-truth annotations and then compute our performance metrics for the Haar Feature-based Cascade Classifier approach by comparing the (corrected) ground truth rectangles to the bounding boxes returned by the detector. For our PED-CED approach we compared detections to ground truth pixel-wise, making the comparison stricter for our approach.
\begin{table}[htb]
\centering
\begin{tabular}{@{}lll@{}}
\toprule
 & PED-CED (Ours) & Haar-based~\cite{a33} \\ \midrule
Accuracy [\%]   	& $99.21\pm0.58$   &  $98.76\pm1.13$  \\
IoU [\%] 	 	& $48.31\pm23.01$ &  $27.23\pm36.47$ \\
Precision [\%]  	& $60.83\pm25.97$ & $36.67\pm46.60$ \\
Recall [\%] 	& $75.86\pm33.11$ & $28.51\pm38.40$ \\
\bottomrule
\end{tabular}
\caption{Comparison of our PED-CED approach and the state-of-the-art Haar-based approach from~\cite{a33}: the average accuracy of a detection prediction with the standard deviation (Accuracy), the Intersection Over Union with the standard deviation (IoU), and in the last two columns Precision and Recall, both with standard deviation values. All values are computed over all 250 test images. Note that Haar-based approach was evaluated using bounded rectangles, whereas our PED-CED approach was evaluated more strictly using pixel-wise comparison. Haar-based approach returnes rectangles as detection which makes pixel-wise comparison impossible to do.}
\label{table_results}
\end{table}

\begin{figure*}[htb]
\captionsetup{type=figure}
	\center
		\subfloat[PED-CED: Accuracy]{\makebox[0.245\columnwidth][c]{
			\includegraphics[width=0.245\columnwidth]{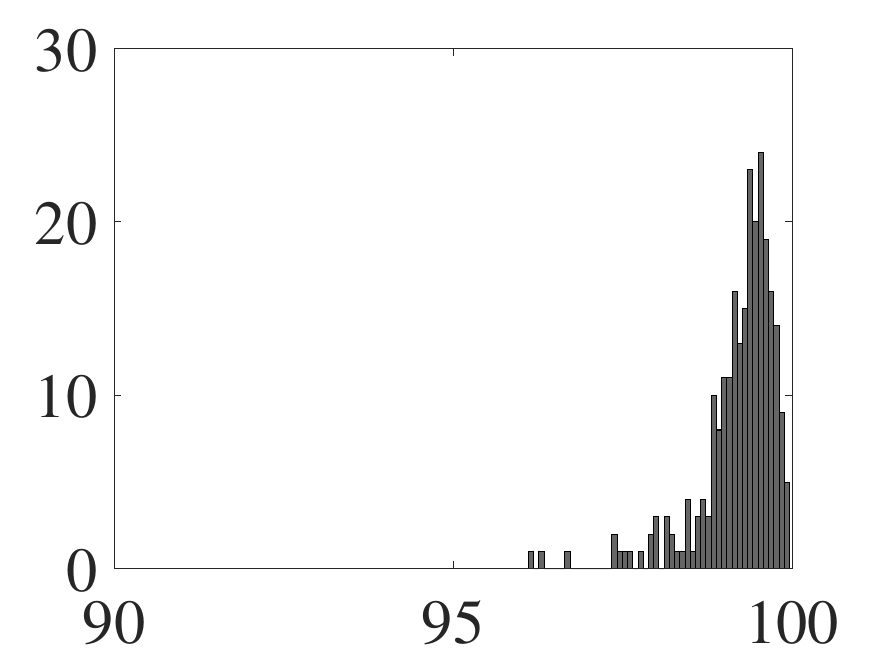}
			\label{img:accuracyHist1}
		}}\hfill
		\subfloat[PED-CED: IoU]{\makebox[0.245\columnwidth][c]{
			\includegraphics[width=0.245\columnwidth]{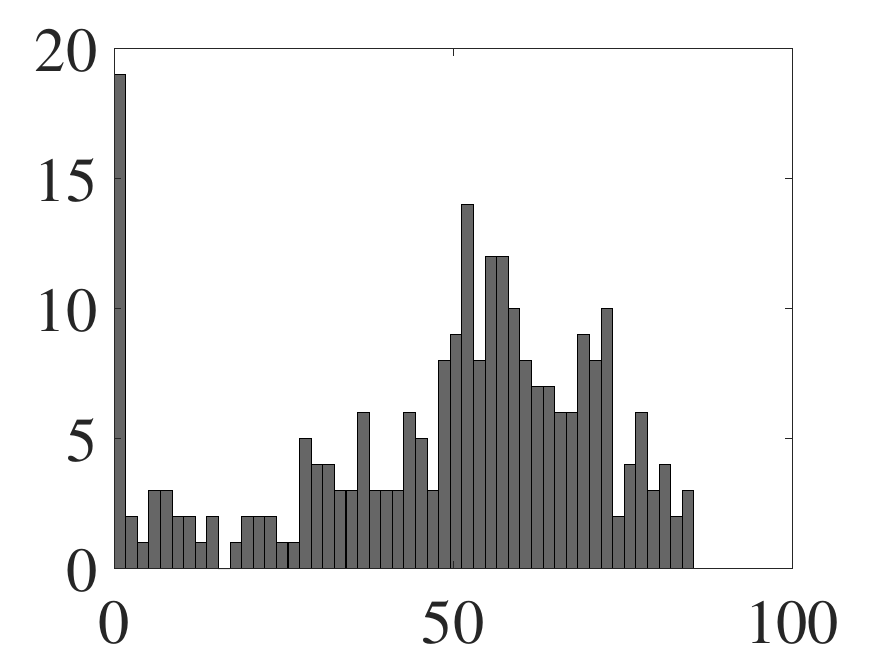}
			\label{img:iouHist1}
		}}\hfill
			\subfloat[PED-CED: Precision]{\makebox[0.245\columnwidth][c]{
			\includegraphics[width=0.245\columnwidth]{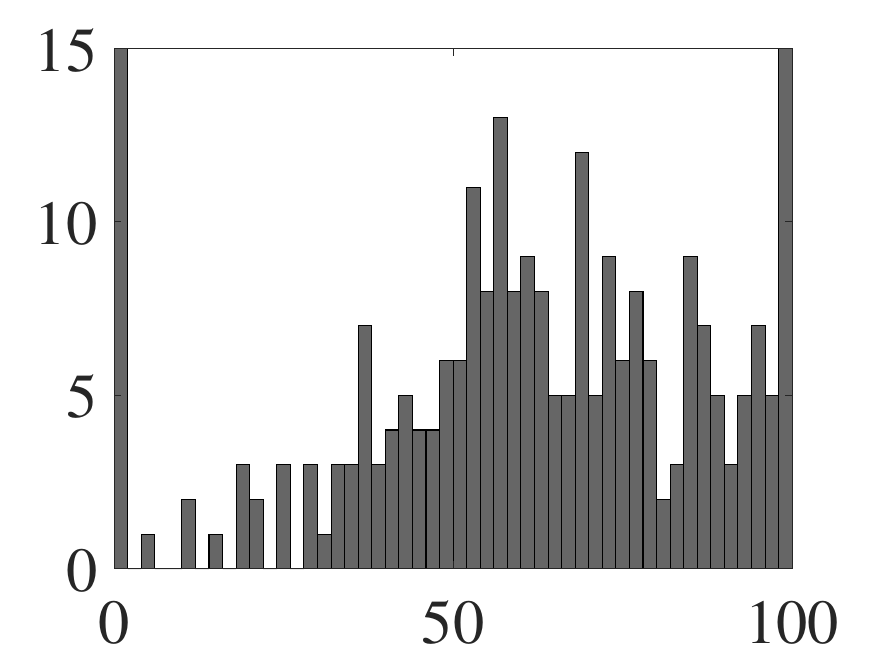}
			\label{img:precHist1}
		}}\hfill
			\subfloat[PED-CED: Recall]{\makebox[0.245\columnwidth][c]{
			\includegraphics[width=0.245\columnwidth]{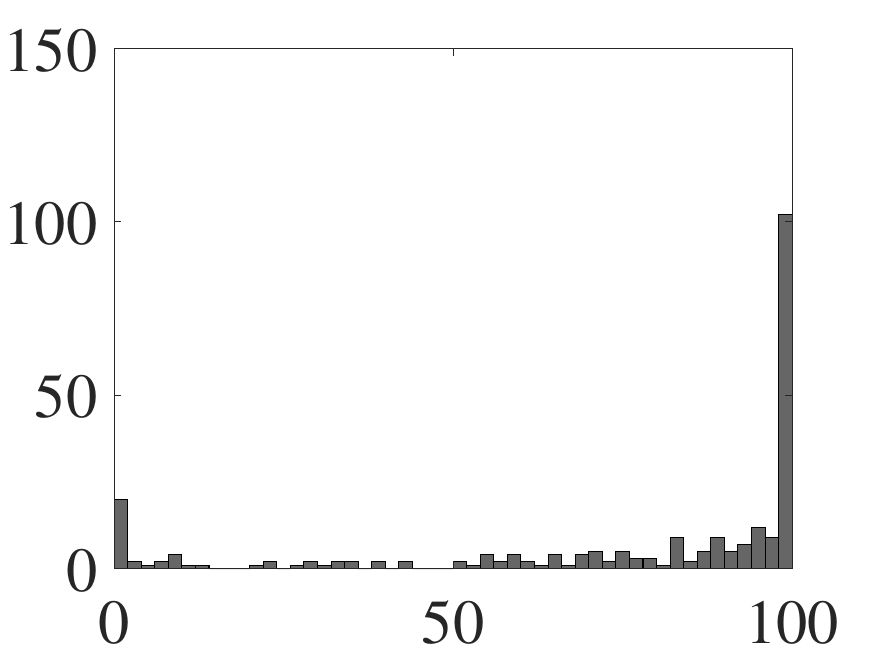}
			\label{img:recallHist1}
		}}\\
\subfloat[Haar: Acc.]{\makebox[0.245\columnwidth][c]{
			\includegraphics[width=0.245\columnwidth]{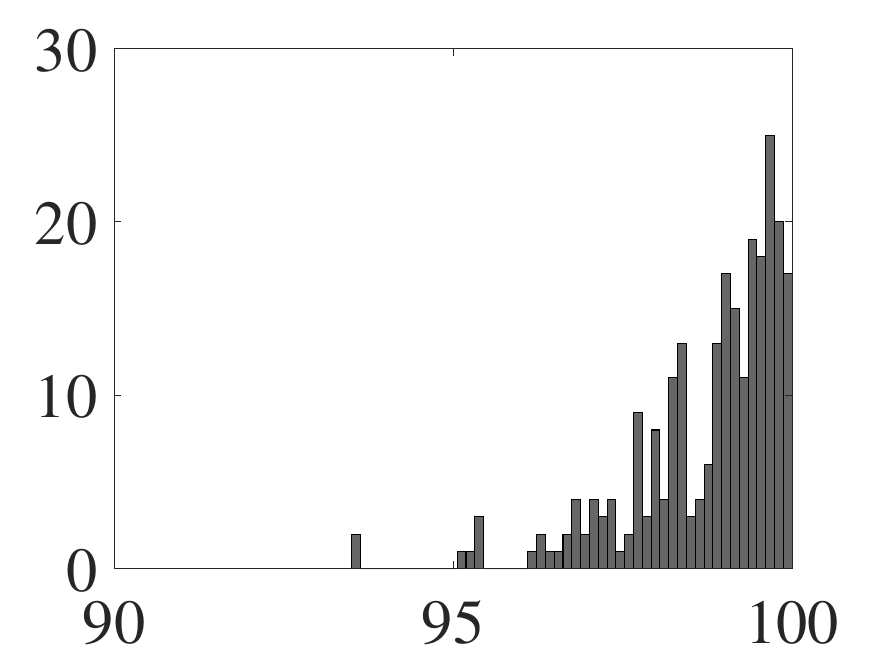}
			\label{img:accuracyHist2}
		}}\hfill
		\subfloat[Haar: IoU]{\makebox[0.245\columnwidth][c]{
			\includegraphics[width=0.245\columnwidth]{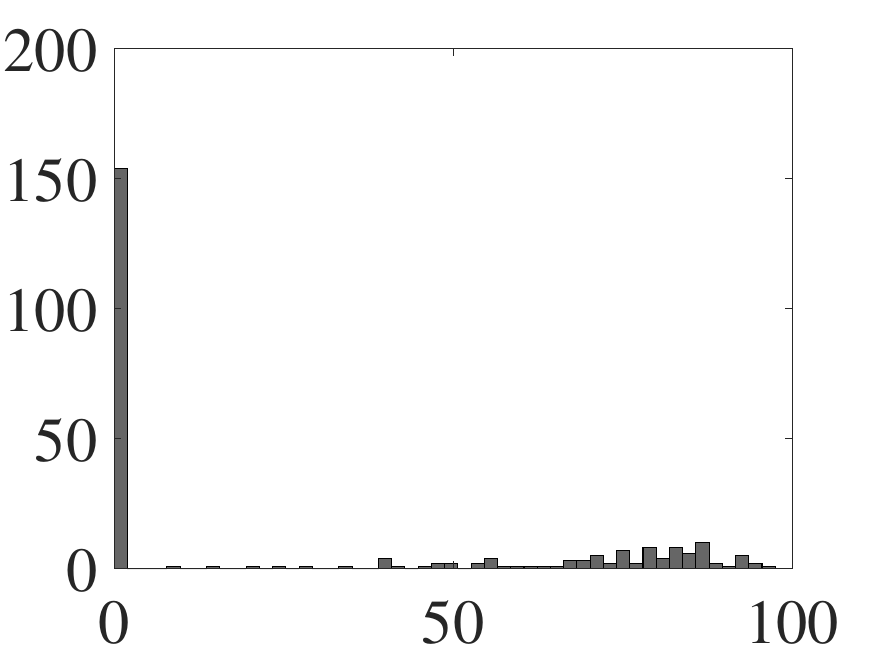}
			\label{img:iouHist2}
		}}\hfill
			\subfloat[Haar: Prec.]{\makebox[0.245\columnwidth][c]{
			\includegraphics[width=0.245\columnwidth]{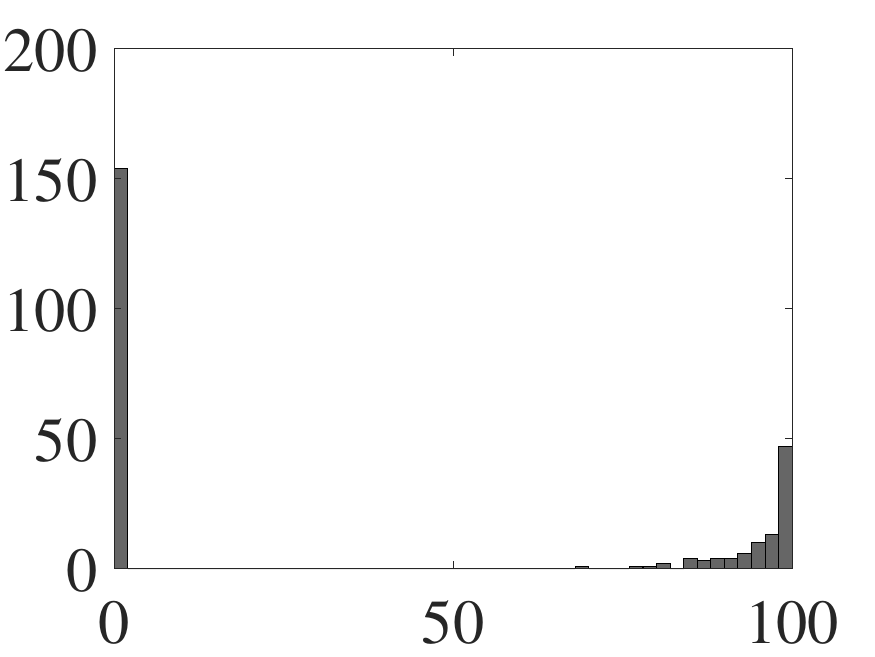}
			\label{img:precHist2}
		}}\hfill
			\subfloat[Haar: Recall]{\makebox[0.245\columnwidth][c]{
			\includegraphics[width=0.245\columnwidth]{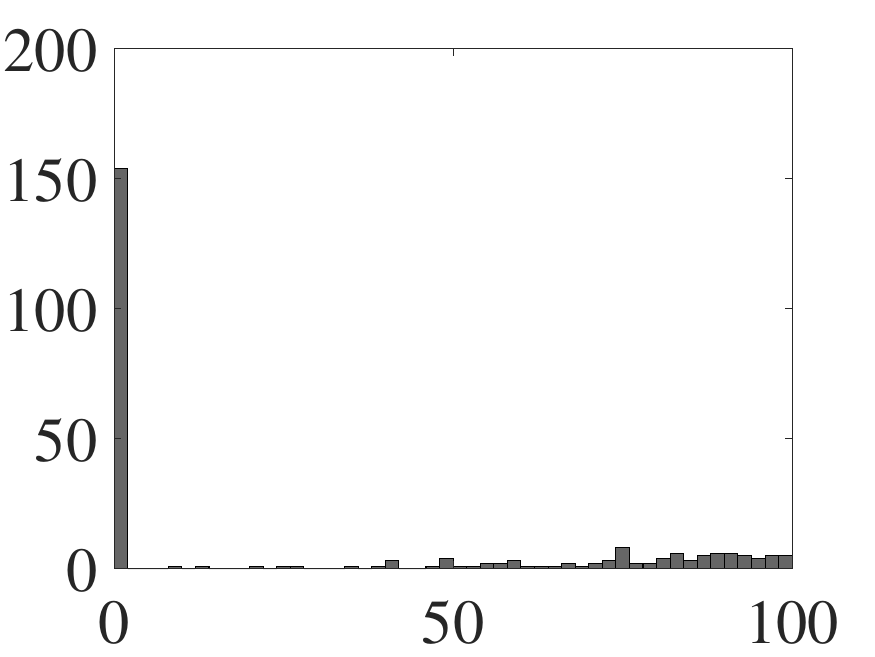}
			\label{img:recallHist2}
		}}
		\caption{Histograms for Accuracy, Intersection over Union (IoU), Precision and Recall for PED-CED approach and Haar Feature-based Cascade Classifier approach. Histograms for PED-CED (the upper row) are significantly better than the Haar Feature-based Cascade Classifier features approach, where the majority of samples is at the 0\%, as opposed ot PEC-CED where samples are mosty evenly distributed or peak at 100\%.}
		\label{img:histograms}
\end{figure*}

The accuracy measure shows how many pixels are correctly classified in the whole image. However, as pointed out in Section~\ref{Sec: perf metrics}, the ear and non-ear class are not balanced (there is significantly more pixels belonging to the non-ear class (the majority class) than to the ear class (the minority class), so we need to be careful when interpreting these results. For our test data, the majority class covers 98.92\% of all pixels and for the minority class cover the remaining 1.08\%. This means that a classifier/detector assigning all image pixels over the entire test set would show an overall accuracy of 98.92\%. Our PED-CNN detection approach achieves the average accuracy of $99.21\%$ whereas the Haar-base detector achieves the accuracy of $98.76\%$
\begin{figure*}[htb]
\captionsetup{type=figure}
	\center
		\subfloat[Head Pitch]{\makebox[0.32\textwidth][c]{
			\includegraphics[width=0.32\textwidth]{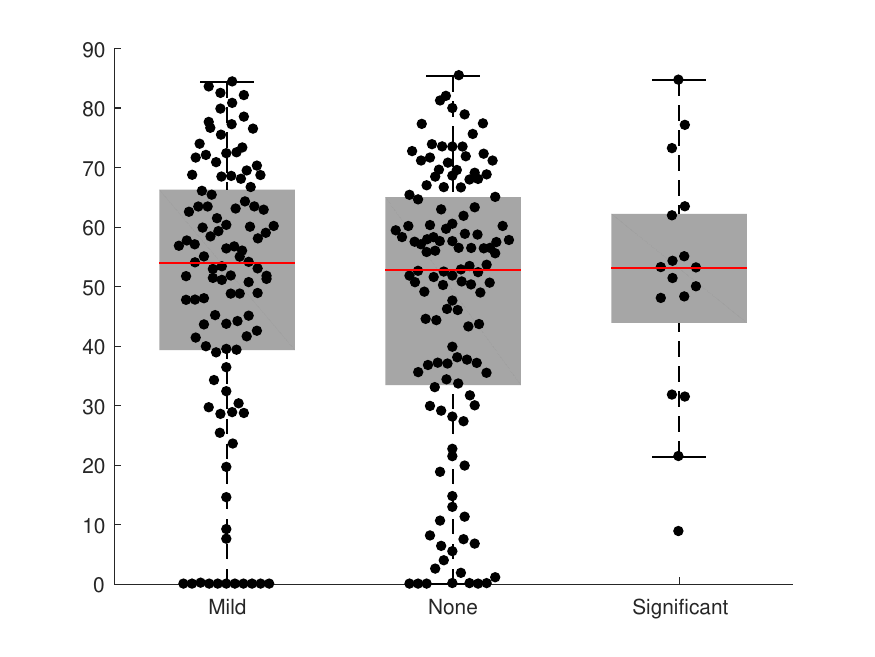}
			\label{img:boxPitch}
		}}\hfill
		\subfloat[Head Roll]{\makebox[0.32\textwidth][c]{
			\includegraphics[width=0.32\textwidth]{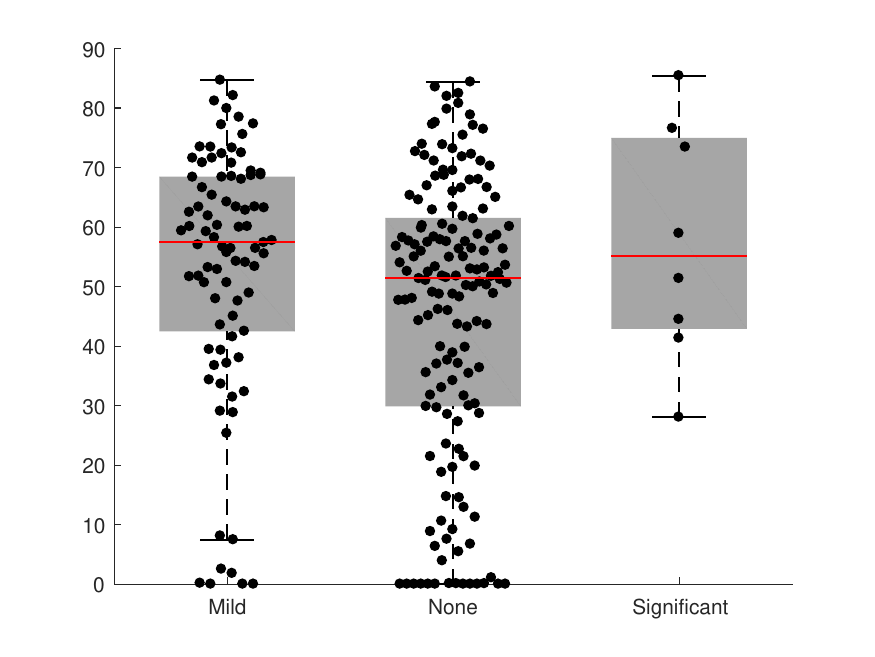}
			\label{img:boxRoll}
		}}\hfill
			\subfloat[Head Yaw]{\makebox[0.32\textwidth][c]{
			\includegraphics[width=0.32\textwidth]{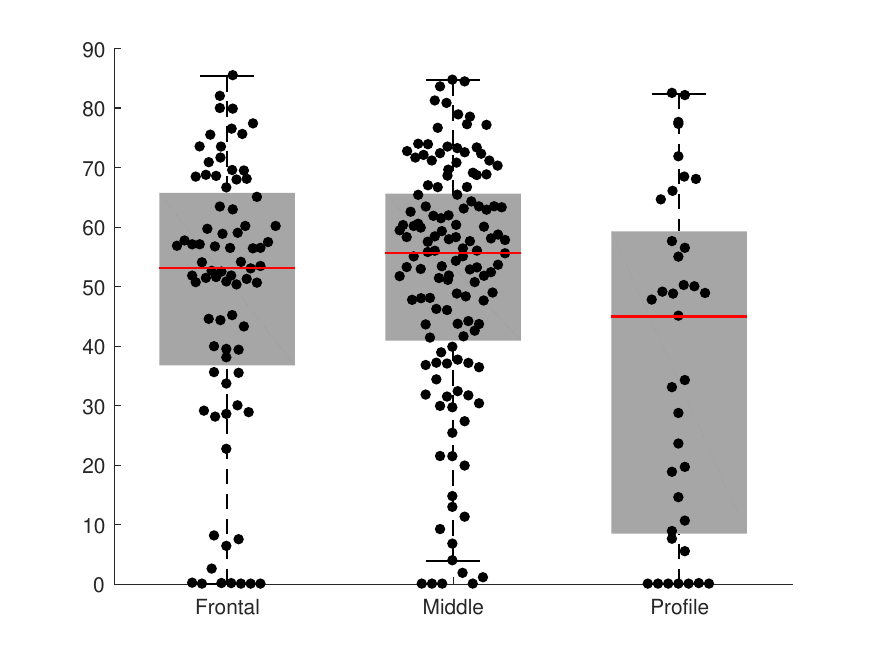}
			\label{img:boxYaw}
		}}\\
			\subfloat[Occlusions]{\makebox[0.32\textwidth][c]{
			\includegraphics[width=0.32\textwidth]{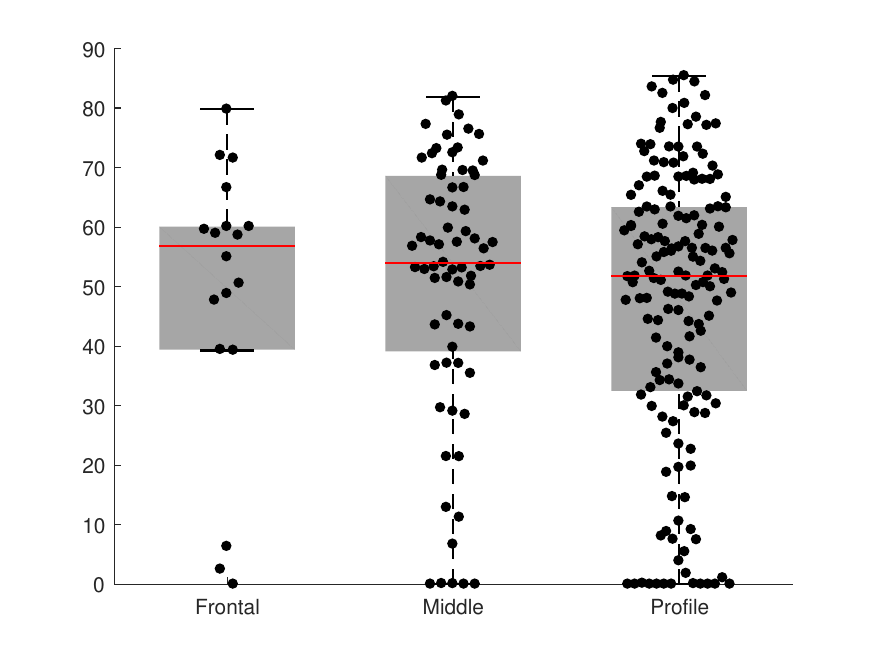}
			\label{img:boxOverlap}
		}}\hfill
			\subfloat[Gender]{\makebox[0.32\textwidth][c]{
			\includegraphics[width=0.32\textwidth]{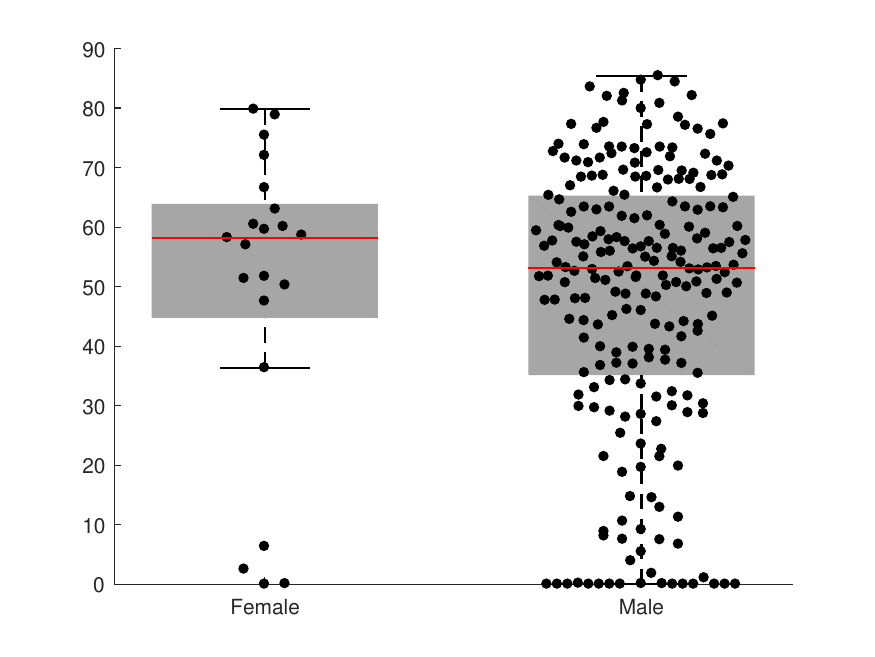}
			\label{img:boxGender}
		}}\hfill
			\subfloat[Ethnicity]{\makebox[0.32\textwidth][c]{
			\includegraphics[width=0.32\textwidth]{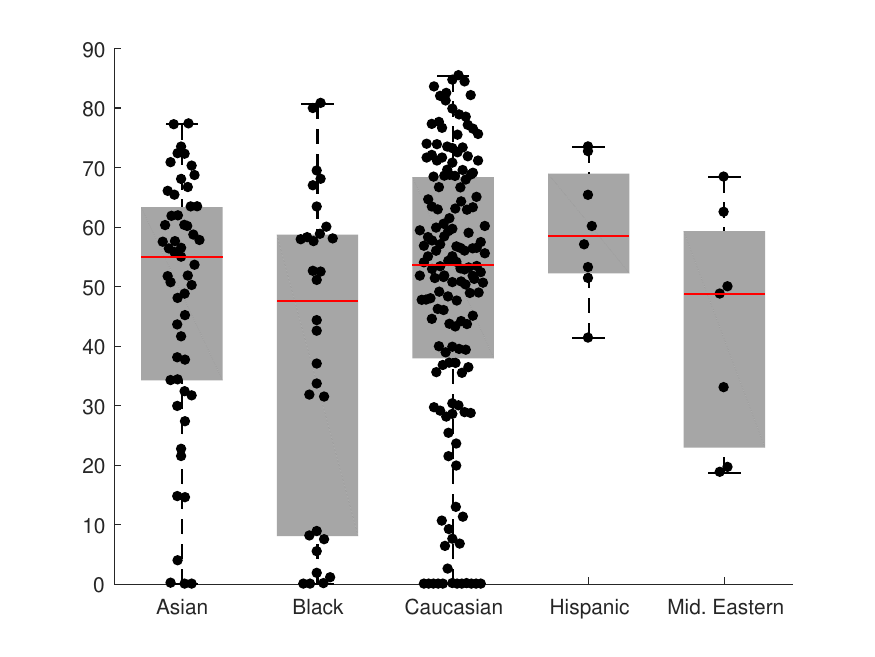}
			\label{img:boxEthnicity}
		}}
		\caption{Impact of various covariates on the performance of our PED-CED ear detection technique. The boxplots show the distribution of IoU values computed over the corresponding test set. Actual IoU values (data points) are superimposed over the boxplots and are shown as black dots. The results show that there is no significant difference in performance across the different covariates, indicating that our approach is robust to various soures of image variability.}
		\label{img:boxplots}
\end{figure*}

The Intersection over Union (IoU) better reflects the actual performance of the evaluated detectors, and is not affected by the distribution of pixels among the majority and minority classes. Every false-positive pixel or true-negative pixel impact the IoU measure significantly. The average IoU for our PED-CED detection approach is $48.31\%$, whereas the Haar-based detector achieves an average IoU of $27.23\%$. The high standard deviations can be attributed to some cases where the detector completely misses the ears. This could be improved by using more diverse images during training, as detection errors typically happen on images with bad illumination, extreme viewing angles and in the presence of severe occlusions (see Section~\ref{Sec: qualitative examples}).

The precision measure is $60.83\pm25.97\%$ for our PED-CED detector and $36.67\pm46.60\%$ for the Haar-based detector. The differences is significant and the high standard deviation in the second case suggests the a high number of complete detection failures for the Haar-based approach. The difference in the recall values is even larger, $75.86\pm33.11\%$ for PED-CED and $28.51\pm38.40\%$ for the Haar-based detector.

Next to the single scalar values presented in Table~\ref{table_results}, we show the complete distribution of the computed performance metrics in Figure~\ref{img:histograms} in the form of histograms. The first column (histograms~\ref{img:accuracyHist1} and~\ref{img:accuracyHist2} shows the distribution of the accuracy measure for the two detection techniques, the second shows the distribution for IoU, the third for precision and the last column shows the distribution of recall values for the entire test set. We see that for the accuracy most of the mass of the distribution of the PED-CED approach is at higher accuracy values than this is the case for the Haar-based approach, which has a heavier tail. In the IoU graphs we observe that our PED-CED approach fails in 19 cases, where the IoU is below $1,5\%$; However, the majority of detections is still between 50\% and 70\% IoU. The Haar-based detector, on the other hand, exhibits a significant peak at the low IoU values with more than 150 images showing an IoU below 5\%. The precision and recall measures show even greater differences. The precision histogram for our PED-CED approach in distributed from 0 to 100\%, whereas the precision histogram for the Haar-based technique is concentrated around 0 and some samples close to 100\% with very little samples in between. This trend is repeated for recall: the majority of values for PEC-CED is around 100\%, whereas the majority of values for Haar is around 0\%.

These results show that the proposed PED-CED detector is able to outperform the state-of-the-art Haar-based approach, but also that the AWE dataset is very challenging, since the Haar-based detector was demonstrated to work well for the ear detection task on a another more controlled dataset~\cite{a33}.

\subsection{Impact of Covariates}

Next, we evaluate the impact of various covariate factors on the performance of our PED-CED ear detection approach. We use the existing annotations that ship with the AWE dataset and explore the impact of: \textit{i)} head pitch, \textit{ii)} head roll, \textit{iii)} head yaw, \textit{iv)} presence of occlusions, \textit{v)} gender, and \textit{vi)} ethnicity. It needs to be noted that the studied covariates are not necessarily unique to each image, so effects of covariate-cross talk may be included in the results. 

We report results in the form of covariate-specific boxplots computed from IoU values in Figure~\ref{img:boxplots}. Here, the actual data points (i.e., IoU values) are also superimposed over the boxplots, which is important for the interpretation of the results, as some covariate classes (i.e., significant head roll, hispanic or middle eastern ethnicity, etc) contain only a few samples. The presented results show that the median value (red lines in the boxplots) of IoU is reasonably close for all classes (or labels) of a given covariate factor and no significant performance difference can be observed for any of the studied covariates. This observation suggests that the propposed PED-CED approach is robust to various sources of variability end ensures stable detection performance in the presence of various sources of image variability.

\subsection{Qualitative evaluation}~\label{Sec: qualitative examples}

\begin{figure*}[t]
\captionsetup{type=figure}
	\center
		\subfloat[85.38\%]{\makebox[0.2\textwidth][c]{
			\includegraphics[width=0.1822\textwidth]{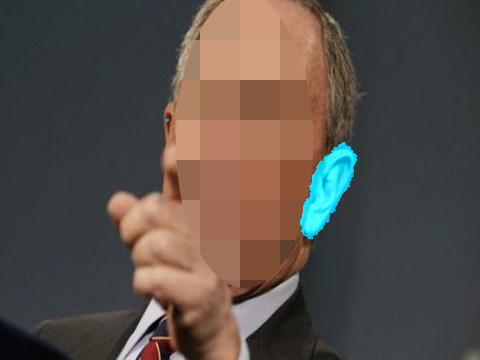}
			\label{img:best1}
		}}\hfill
		\subfloat[84.63\%]{\makebox[0.2\textwidth][c]{
			\includegraphics[width=0.1822\textwidth]{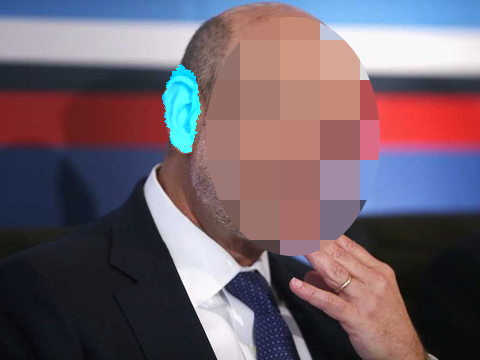}
			\label{img:best2}
		}}\hfill
		\subfloat[83.49\%]{\makebox[0.2\textwidth][c]{
			\includegraphics[width=0.1822\textwidth]{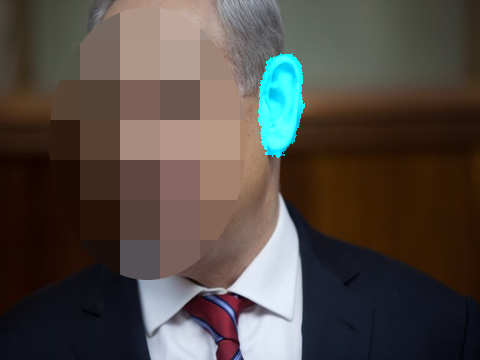}
			\label{img:best3}
		}}\hfill
		\subfloat[82.41\%]{\makebox[0.2\textwidth][c]{
			\includegraphics[width=0.1822\textwidth]{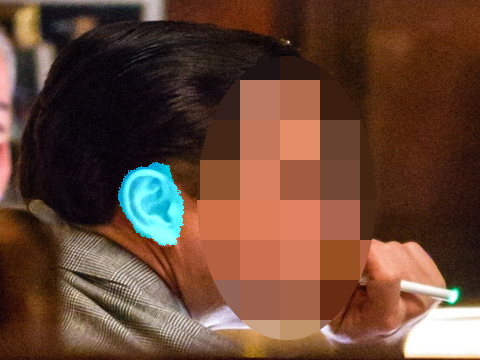}
			\label{img:best4}
		}}\hfill
		\subfloat[79.86\%]{\makebox[0.2\textwidth][c]{
			\includegraphics[width=0.1822\textwidth]{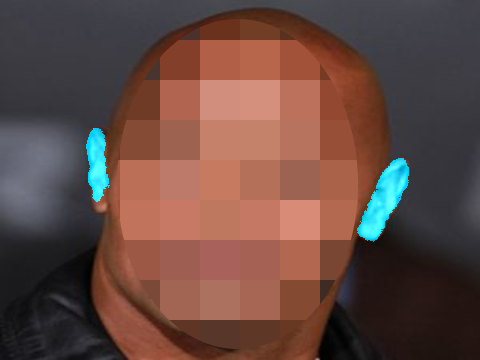}
			\label{img:best5}
		}}%
		\\ \vspace{-1ex}%
		\subfloat[51.00\%]{\makebox[0.2\textwidth][c]{
			\includegraphics[width=0.1822\textwidth]{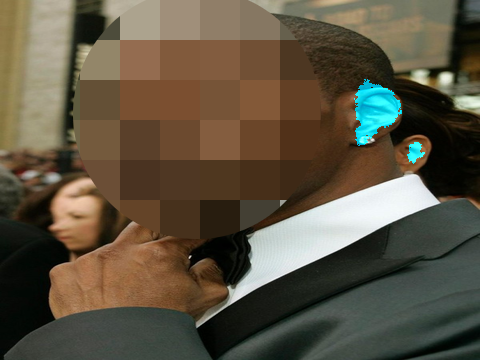}
			\label{img:middle1}
		}}\hfill
		\subfloat[50.74\%]{\makebox[0.2\textwidth][c]{
			\includegraphics[width=0.1822\textwidth]{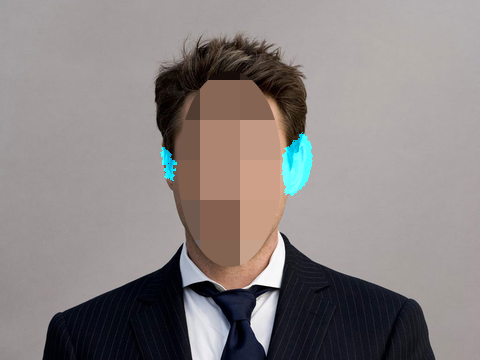}
			\label{img:middle2}
		}}\hfill
		\subfloat[50.63\%]{\makebox[0.2\textwidth][c]{
			\includegraphics[width=0.1822\textwidth]{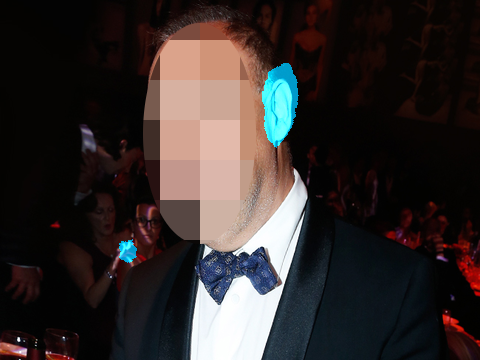}
			\label{img:middle3}
		}}\hfill
		\subfloat[50.24\%]{\makebox[0.2\textwidth][c]{
			\includegraphics[width=0.1822\textwidth]{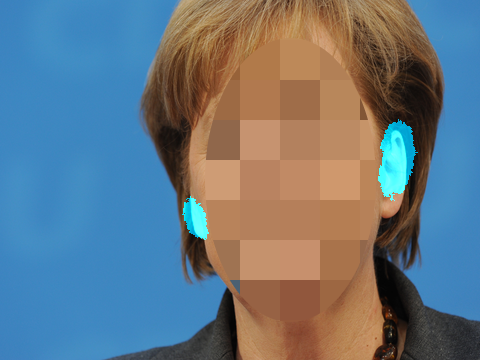}
			\label{img:middle4}
		}}\hfill
		\subfloat[49.02\%]{\makebox[0.2\textwidth][c]{
			\includegraphics[width=0.1822\textwidth]{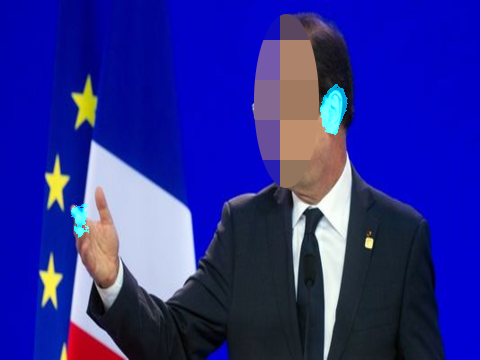}
			\label{img:middle5}
				}}%
		\\ \vspace{-1ex}%
		\subfloat[0.10\%]{\makebox[0.2\textwidth][c]{
			\includegraphics[width=0.1822\textwidth]{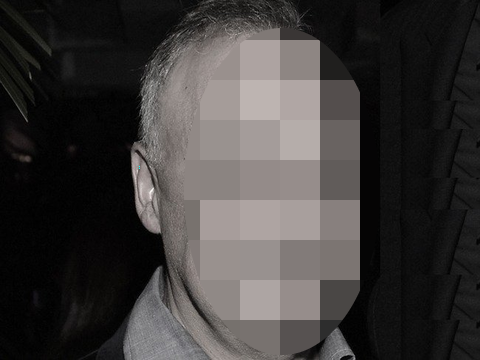}
			\label{img:worst1}
		}}\hfill
		\subfloat[0.08\%]{\makebox[0.2\textwidth][c]{
			\includegraphics[width=0.1822\textwidth]{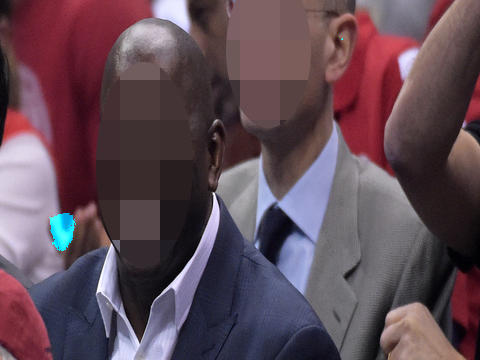}
			\label{img:worst2}
		}}\hfill
		\subfloat[0.08\%]{\makebox[0.2\textwidth][c]{
			\includegraphics[width=0.1822\textwidth]{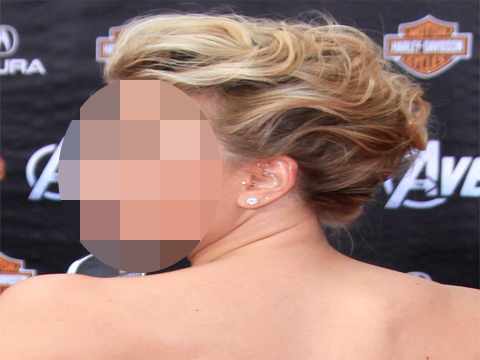}
			\label{img:worst3}
		}}\hfill
		\subfloat[0\%]{\makebox[0.2\textwidth][c]{
			\includegraphics[width=0.1822\textwidth]{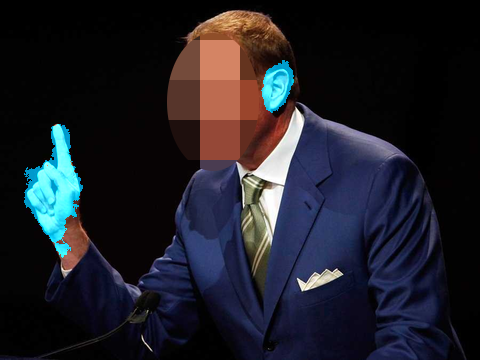}
			\label{img:worst4}
		}}\hfill
		\subfloat[0\%]{\makebox[0.2\textwidth][c]{
			\includegraphics[width=0.1822\textwidth]{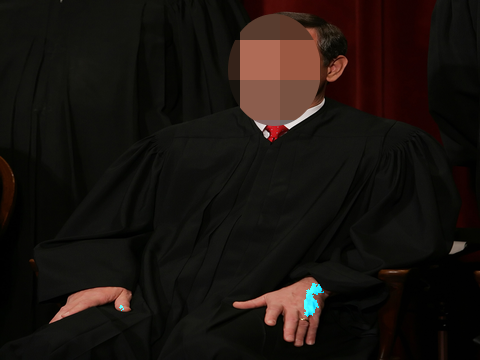}
			\label{img:worst5}
		}}

		\caption{Examples of our detection results ordered in terms of decreasing values of IoU. The top row shows some of the best detections, the middle row shows average detections (note that the mean value of IoU for our entire test set was $48.31\%$) and the last row shows some of the worst detection examples. All listed percentage values represent IoU-s.}
		\label{img:detectionRepresentatives}
\end{figure*}

Last but not least, we show a few qualitative examples of our detection results in Figure~\ref{img:detectionRepresentatives}. The first row of images show the best detections with IoU values above $80\%$, the second row shows detections with IoU values around $50\%$, which is also close to the average value achieved on our test set (see Table~\ref{table_results}) and the last row shows the worst detection results with IoU values close to 0\%. These last examples represent examples of the 19 failed cases from our test set.

It needs to be noted that the IoU values around $50\%$ (middle row in Figure~\ref{img:detectionRepresentatives}) are achieved on difficult images with variations across pose, race, occlusion and so forth. These values are also more than sufficient to ensure good detections that can be exploited by fully automatic recognition systems. On the other hand, detections with IoU values close to $0\%$ are of no use to biometric systems and represent cases where our method is in need of improvement. These cases occur with images captured at extreme viewing angles with respect to the ears and in images with limited contrast among others.

\section{Conclusion}

Ear detection in unconstrained conditions is a difficult problem  affected by: different angles from which images are taken, various skin color tones, changes illumination conditions, occlusions, accessories. In order to address the problem of ear detection in unconstrained environments successfully we proposed in this work a new ear detector based on convolutional encoder-decoder network.

Our experiments showed that our approach detects ears with the average accuracy of 99.21\%, an average Intersection over Union (IoU) value of 48.31\%, average precision of 60.83\% and the average recall of 75.86\%. All of these performance metrics were also shown to be significantly higher than those achieved by the competing state-of-the-art Haar-based ear detector.

Our future work with respect to ear detection includes incorporating contextual information into the detection pipeline, looking for ears only in the vicinity of faces and in specific relation to other facial parts (such, as the nose, eyes, etc.).

Our long-term plan for the future is to incorporate this presented detection method into a pipeline of ear recognition. The system like that will be able to recognize persons based on ears only by inputting plain images taken in-the-wild. Furthermore, considering the speed of the current implementation of our ear detector, such an approach should be able to operate in real-time or close to real-time at the very least.

\section*{Acknowledgements}

This research was supported in parts by the ARRS (Slovenian Research Agency) Research Programme P2-0250 (B) Metrology and Biometric Systems as well as the ARRS Research Programme P2-0214 (A) Computer Vision.  

\bibliographystyle{IEEEtran}
\bibliography{main}

\end{document}